\documentclass[conference]{IEEEtran}
\IEEEoverridecommandlockouts
\usepackage{cite}
\usepackage{amsmath,amssymb,amsfonts}
\usepackage{algorithmic}
\usepackage{graphicx}
\usepackage{textcomp}
\usepackage{xcolor}
\usepackage{url}
\usepackage{tikz}
\def\BibTeX{{\rm B\kern-.05em{\sc i\kern-.025em b}\kern-.08em
    T\kern-.1667em\lower.7ex\hbox{E}\kern-.125emX}}
\begin{document}

\title{Exascale Deep Learning to Accelerate Cancer Research \\
\thanks{This manuscript has been authored by UT-Battelle, LLC under Contract
No. DE-AC05-00OR22725 with the U.S. Department of Energy.  The United
States Government retains and the publisher, by accepting the article
for publication, acknowledges that the United States Government
retains a non-exclusive, paid-up, irrevocable, worldwide license to
publish or reproduce the published form of this manuscript, or allow
others to do so, for United States Government purposes. The
Department of Energy will provide public access to these results of
federally sponsored research in accordance with the DOE Public Access
Plan (http://energy.gov/downloads/doe-public-access-plan).}
}

\author{

\IEEEauthorblockN{Robert M. Patton, J. Travis Johnston, Steven R. Young, Catherine D. Schuman, Thomas E. Potok, \\ Derek C. Rose, Seung-Hwan Lim, Junghoon Chae}
\IEEEauthorblockA{ \textit{Oak Ridge National Lab} \\
Oak Ridge, TN, United States \\
\{pattonrm, johnstonjt, youngsr, schumancd, potokte, rosedc, lims1, chaej\}@ornl.gov}

\and

\IEEEauthorblockN{Le Hou, Shahira Abousamra, Dimitris Samaras}
\IEEEauthorblockA{ \textit{Department of Computer Science} \\
\textit{Stony Brook University} \\
Stony Brook, NY, United States \\
le.hou@stonybrook.edu, \{sabousamra, samaras\}@cs.stonybrook.edu}

\and

\IEEEauthorblockN{Joel Saltz}
\IEEEauthorblockA{\textit{Department of Biomedical Informatics} \\
\textit{Stony Brook University} \\
Stony Brook, NY, United States \\
joel.saltz@stonybrookmedicine.edu}
}
\maketitle

\begin{abstract}
Deep learning, through the use of neural networks, has demonstrated remarkable
ability to automate many routine tasks when presented with sufficient data
for training.  
The neural network architecture (e.g. number of layers, types of layers, connections between layers, etc.) plays a critical role in determining what, if
anything, the neural network is able to learn from the training data.
The trend for neural network architectures, especially those trained on ImageNet, has been to grow ever deeper and more complex.
The result has been ever increasing accuracy on benchmark datasets with the cost of increased computational demands.
In this paper we demonstrate that neural network architectures can be automatically generated, tailored for a specific application, with dual objectives: accuracy of prediction and speed of prediction.
Using MENNDL--an HPC-enabled software stack for neural architecture search--we generate a neural network with comparable accuracy to state-of-the-art networks on a cancer pathology dataset that is also $16\times$ faster at inference.
The speedup in inference is necessary because of the volume and velocity of cancer pathology data; specifically, the previous state-of-the-art networks are too slow for individual researchers without access to HPC systems to keep pace with the rate of data generation.  Our new model enables researchers with modest computational resources to analyze newly generated data faster than it is collected.
\end{abstract}


\section{Introduction}

Many scientific research applications rely on the analysis of imagery produced using specialized instruments such as microscopes.  Scientific data analysis can be accelerated through effective use of automation techniques such as deep learning.  There are two obstacles for effective deployment of deep learning for scientific data analysis.  First, scientific data does not often share characteristics with standard datasets (such as ImageNet).  As such, deep learning models optimized for standard datasets are not always ideal for novel scientific datasets.  Second, the volume and velocity of scientific data production necessitate models that can make very fast and accurate predictions.  Because of these unique challenges, tailoring neural network architectures for a specific dataset is often necessary.  Even with human guided design, designing deep networks for fast and accurate predictions can be computationally expensive and time-consuming.

In this work, we demonstrate a software framework called Multinode Evolutionary Neural Networks for Deep Learning (MENNDL) that utilizes high-performance computing (HPC) to automate the design of deep learning networks in order to analyze cancer pathology images.  MENNDL effectively parallelizes and scales the evaluation of millions of networks within hours, utilizing the computational power of a GPU-based HPC system.  Here, we have enhanced MENNDL to allow for multi-objective optimization, so that factors such as prediction time are explicitly included as objectives during the network design process.  We demonstrate that MENNDL can produce a sufficiently accurate model for a cancer pathology task that makes predictions $16\times$ faster than state-of-the-art approaches.  The reduced prediction time allows for more practical deployment of deep learning to analyze both existing cancer pathology data and new data as it is collected.  We also benchmark MENNDL showing that it can achieve 1.3 Exaflops (mixed precision floating point operations) on Oak Ridge National Laboratory's Summit supercomputer--currently, the fastest supercomputer in the world.   

The key contributions of this work are:
\begin{itemize}
    \item A 16$\times$ performance improvement in the rate of TIL classification within whole slide pathology images, achieved by utilizing multi-objective MENNDL on Summit at scale.  This capability enables the cancer research community to analyze pathology data at an unprecedented scale.
    \item A deep learning software system, MENNDL, that can use HPC to optimize the network design and hyperparameters for high accuracy, as well as additional objectives that influence real-world performance, such as shorter prediction times or lower energy for prediction. 
    \item The first deep learning system operating at over 1.3 exaflops, achieved by exploiting the tensor cores on the NVIDIA GPUs on the Summit supercomputer.
    \end{itemize}

\section{Background and Related Work}

Our contributions in this work are in three areas: 1) automatically annotating digital pathology images for cancer research, 2) multi-objective optimization in deep learning, and 3) effectively utilizing HPC systems for deep learning. Thus, we discuss the current state-of-the-art in each of those fields.

\subsection{Tumor Infiltrating Lymphocytes Classification}

The target application in this work is annotating cancer pathology data. During the cancer diagnosis and treatment process, a patient may have a biopsy, which produces a diagnostic tissue sample. Using this sample, a slide is prepared and examined under a microscope by a pathologist to understand both how to treat the disease and to provide a prognosis for the patient's future. Virtually all cancer patients undergo these biopsies, producing large volumes of these pathology slides.  

Digital pathology, or the review of digitized pathology slides, is gaining traction in part because quantitative measurements on digitized whole slide images lead to reproducible and significantly nuanced observations. The recent FDA approval \cite{FDANewsRelease} of whole slide imaging for primary diagnostic use is leading to widespread adoption of digital whole slide imaging. It is widely expected that within 5-10 years the great majority of new pathology slides will be digitized.  A significant feature in these images is tumor infiltrating lymphocytes (TILs), which are types of immune cells that move into a tumor to try to attack the cancer. The quantification of TILs is well known to have prognostic value in many contexts \cite{Fridman2012}\cite{ANGELL2013261} because understanding patient immune response to tumors is becoming increasingly important with the growth of cancer immunotherapy. Features such as TILs can be quantified through image analysis and deep learning algorithms \cite{saltz2018spatial},\cite{klauschen2018scoring}. A whole slide image will typically contain hundreds of thousands of labeled regions, each approximately $50 \times 50$~$\mu$m. Each resulting image is typically gigapixel in scale and encompasses 100,000 to 1,000,000 cell nuclei.

The clinical and research demand for TIL quantification is rapidly growing \cite{savas2016clinical}. TIL characterization is proving to be clinically predictive in many disease sites. Understanding the presence of TILs on a population level can provide a deeper mechanistic understanding of the role of intra-tumoral immunity in cancer progression and treatment.    Pathologists often use only high level terms such as ``brisk" or ``sparse" to describe TILs. There is a strong ongoing effort to train pathologists to provide manual estimates of TIL percentages; however, quantitative estimates of TIL coverage in tumors are subject to high inter-observer variability \cite{swisher2016interobserver}. The description of TIL spatial patterns is even more problematic. Manual demarcation of TIL rich regions is very time consuming and rarely carried out except on a very small number of specimens in research settings.    

In \cite{saltz2018spatial}, a variety of deep learning methods were developed that were able to provide quantitative estimates of TIL density as well as to characterize patterns of TIL distribution. The quantitative TIL characterization goes far beyond what a pathologist can estimate by inspecting tissue.  In that work, a convolutional neural network (CNN) architecture is systematically optimized to carry out recognition and classification of nuclei from pathology images.  This led to the release of a dataset consisting of TIL maps corresponding to roughly 5,000 whole slide images from The Cancer Genome Atlas (TCGA). The work also encompassed a thorough set of analyses that related TIL spatial patterns to a rich set of molecular immune and tumor features.  Slow prediction time is, however, a serious drawback to the use of either the standard or hand-crafted CNN architectures \cite{thorsson2018immune}.  

\subsection{Multi-Objective Optimization in Deep Learning}
Within the deep learning community, focus has primarily been directed at achieving higher accuracy or lower mean squared error for given tasks, while neglecting other factors important to deployment, such as prediction time.  Evolutionary optimization has been leveraged in order to both design the network topology and optimize the weights in the network \cite{Montana:1989:TFN:1623755.1623876}\cite{273950}\cite{Stanley:2002:ENN:638553.638554}. Building upon these prior works \cite{Roth2006} \cite{7920404} we extend our evolutionary optimization algorithm to include multiple objectives.  In \cite{Roth2006}, their primary objectives are 1) sparsity of the network and 2) classification accuracy.  Sparse networks typically generalize patterns better (giving higher accuracy) and theoretically run faster. However, the authors note that these two objectives may be conflicting.  In recent work, the most accurate networks have often been deeper, more complex networks that require longer prediction times \cite{DBLP:journals/corr/CanzianiPC16}. At the time the work was performed, \cite{Roth2006} showed that evolutionary optimization outperformed other methods in the task of face detection by achieving sparser networks with lower classification error. Later work confirms that deeper networks are not necessarily better and shows that shallow networks can be made to perform just as well as deeper, more computationally complex networks \cite{NIPS2014_5484}.  Furthermore, while not utilizing multi-objective optimization, the work of \cite{Mocanu2018} produced similar results showing that an evolutionary optimization approach can produce sparse networks with high classification accuracy.

\subsection{Deep Learning \& High Performance Computing}

The use of GPU-based HPC systems is increasingly popular in the field of deep learning applications~\cite{deeplearningHPCsurvey},
partially because they have been successfully applied to a variety of applications and partially because deep learning applications are typically very computationally intensive.  There are three phases of designing and using a deep learning approach for a given dataset, and HPC can be used for any of the three phases.  These phases are 1) design, 2) training, and 3) deployment.

In the first phase, the design of the network topology and hyperparameters are customized for the dataset. The most common approach for this step is for the researcher who is using the deep learning approach to hand-tune the design of the network, typically starting from some other popular network design and performing intuition-guided (slightly less than random) adjustments to a variety of parameters until the accuracy of the network is increased. It is not uncommon for this manual process to take months. Leveraging HPC along with supplementary machine learning techniques has been shown to both reduce the time to solution down to hours, as well as increase the performance of the network, often beyond what a domain expert could achieve \cite{young2017evolving}.  There have been a variety of approaches for designing hyperparameters for deep learning on HPC, including using support vector machines to drive prediction of good hyperparameter sets \cite{Johnston:2017:OCN:3146347.3146352}, using Bayesian optimization \cite{balaprakash2018deephyper}, and using genetic algorithms \cite{young2017evolving, martinez2018deep}. HPC is very well-suited to addressing this phase of deep learning as determining the appropriate design for the deep learning network often requires evaluating thousands to millions of potential network designs. 

In the second phase, the training of the network (i.e., the training of the weights in the network for a particular task) is parallelized in a variety of ways. In one approach, the training data is passed through the network in batches with each node working with different batches of data and sharing weight updates. This approach is called data parallelism but has been shown to have issues with scaling computationally while maintaining algorithmic performance \cite{Kurth:2018:EDL:3291656.3291724}. In another approach, called model parallelism, the network's layers can be distributed across multiple GPUs and/or multiple nodes in order to train the entire network as a whole \cite{VanEssen:2015:LLB:2834892.2834897}. However, this approach is severely limited in its ability to scale, as deep networks that span entire supercomputers have not yet been shown to outperform those networks that can fit on a single node.

In the third and final phase, once a network has been designed and trained, it is then deployed and used to analyze real data, called inferencing or prediction. Utilizing HPC for prediction can be very valuable, particularly when model parallelism is required for very large datasets. However, because most domain scientists do not have access to leadership-class HPC, this is often not a practical use of deep learning on HPC. Though there is a small, limited scope scenario in which HPC would be useful for prediction, for broader impact, using HPC to create a model that can rapidly and accurately classify data on non-HPC systems is a potential breakthrough capability for both the HPC and the scientific community. 
\section{Methods}

\subsection{MENNDL}

Our deep learning framework, MENNDL, relies on an evolutionary algorithm to determine the optimal deep neural network design for a particular scientific dataset \cite{young2017evolving}. MENNDL is designed to leverage GPU-based HPC systems to produce deep learning networks that are optimized for scientific datasets in a quick, efficient, and automated manner. It can be utilized on smaller scale HPC systems (e.g. an 8 GPU system) as well as larger HPC resources (e.g the Summit supercomputer) depending on the complexity of the problem and the desired time to solution. MENNDL optimizes the network topology (number and type of layers), as well as the hyperparameters of each layer in the network in order to customize it for a particular dataset.  This approach is contrasted with the typical approach taken for scientific data, which is to use an ``off-the-shelf" network topology and hyperparameter set and then hand-tune the topology and hyperparameters for the dataset. This hand-tuning process can take on the order of weeks to months for a scientist to perform.  Utilizing MENNDL and a GPU-based HPC system, this process is automated and can take on the order of hours, while typically achieving performance better than the network hand-tuned by a domain expert. In a previous work, we introduced the scalability of the original MENNDL code, demonstrating a peak performance of 167 sustained petaflops on designing a deep neural network for use on scanning transmission electron microscopy data \cite{patton2018167}.  The scalability of MENNDL relies on its asynchronous, master-worker genetic algorithm implementation, which is used to keep as many GPUs as are available busy evaluating candidate network designs over the course of the evolution. 

MENNDL is wrapped around a deep learning implementation framework (e.g., PyTorch, TensorFlow), which is used to evaluate each candidate network topology and hyperparameter set by training a network's weights.  Previous versions of the MENNDL software were written in C++ and utilized the Caffe framework as the deep learning backend \cite{patton2018167}.  This work uses a new version of the MENNDL software, which is written in Python and utilizes PyTorch as the deep learning backend.  Both Caffe and PyTorch utilize cuDNN, NVIDIA's GPU-accelerated library of primitives for deep neural networks.  Message Passing Interface (MPI) is used to communicate between nodes in MENNDL, and the data is stored in a Lightning Memory-Mapped Database (LMDB).

\subsubsection{Multi-Objective MENNDL}

Here, we expand on MENNDL by augmenting the underlying evolutionary algorithm to perform multi-objective optimization.  Previously, MENNDL would evolve towards optimizing the network to maximize accuracy or some other single objective of performance, like F1-score, on a particular task.  Now, we have expanded the evolutionary process to potentially include optimizing the following objectives: 1) minimizing training time, 2) maximizing resource utilization during training, 3) maximizing accuracy on the given dataset, 4) minimizing network size, and/or 5) minimizing prediction time.

All of these objectives are interrelated and can potentially affect each other. Maximizing resource utilization during training is a system-specific objective.  When using Summit, the training of each individual network utilizes one NVIDIA Volta GPU.  In this case, one definition for maximizing resource utilization during training is to maximize the usage of the Volta's tensor cores.  By maximizing a network's utilization of the tensor cores, it is likely that the training time is also reduced.  Training time and prediction time are also highly correlated with network size, simply because smaller networks require fewer operations in both the training and prediction stages.  
  
Previous work with MENNDL uses accuracy on the validation set as the fitness score in the evolutionary algorithm to drive future model refinement and selection.  With multi-objective optimization, we include the additional objectives as part of the fitness function.  In particular, the fitness function $f$ of given network $net$, is defined as follows:

\begin{equation}
    f(net) = v(net) + \alpha m(net)
\end{equation}

\noindent where the function $v$ gives the performance on the validation set of a given network. The performance can be measured in different ways, such as accuracy or F1-score on a validation set.  The function $m$ gives the measurement of a given objective (i.e., training time, network size, etc.), that is normalized to a value between 0 and 1. The $\alpha$ parameter is a scaling factor that can be positive or negative depending on whether the additional objective is to be maximized or minimized. In this work, we restrict our attention to maximizing the F1-score on the validation set and minimizing prediction time.

\subsubsection{Optimization for Faster Training}

It is important to note that the goal of MENNDL is not to produce a fully trained model (i.e., a network that can immediately be deployed on a particular problem), but to optimize the network design to perform well on a particular dataset. The resulting network design can then be trained for a longer period of time on the dataset to produce a fully trained model.  During MENNDL's network design evaluation process, millions of potential networks are evaluated, where each network is trained for a short period of time to assess the performance of that particular network design against a given dataset.  As this is the most computationally intensive part of MENNDL (i.e., the training of each individual network to produce a fitness score), we include a variety of optimizations to speed up the training time for each individual network design.  

\begin{figure}[htb]
    \centering
    \includegraphics[width=\columnwidth]{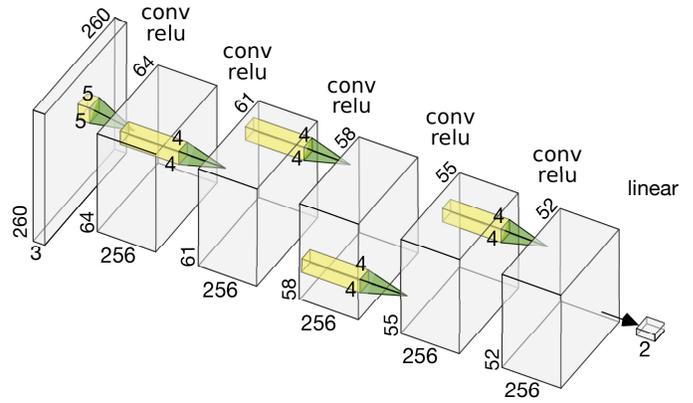}
    \caption{Network that utilizes the high-performing network hyperparameters. Figure created using \cite{lenail2019nn}.}
    \label{fig:wtn}
\end{figure}

Different network hyperparameters and topologies have radically different utilization of the tensor cores in NVIDIA Volta GPUs and thus radically different performance characteristics.  We performed a sweep of approximately 10,000 convolution layer hyperparameter sets to identify which configuration maximally utilizes the tensor cores, which are specialized hardware designed to accelerate deep learning operations.  
Figure \ref{fig:hyperparameter_sweep} highlights the top ten performing networks in terms of GPU utilization or FLOP/s.
As can be seen in this figure, there are certain hyperparameters and associated values that are correlated with a higher number of FLOP/s.  In particular, setting the number of input channels at 256, the number of outputs at 256, the kernel size at 4, and the stride at 1 resulted in the top ten performing networks in terms of FLOP/s, whereas batch size had relatively little impact. 
Using these results, we can then bias network hyperparameters within the MENNDL optimization towards those that have higher utilization of the tensor cores and thus should train faster.  Using these high-performing network hyperparameters, we have also manually constructed a network (shown in Figure \ref{fig:wtn}) that operates on the dataset described in Section \ref{sec:performed_measured:data_description}, which we use to benchmark system performance.

\subsection{Cancer Research Application}
\label{sec:innovations_cancer}

For this work, we focus on using MENNDL to produce a neural network capable of rapidly generating TIL characterizations on commodity hardware. The need to characterize the state of tumor immune interactions is ubiquitous given the pivotal role played by the immune system in cancer. The World Health Organization (WHO) International Agency for Research on Cancer estimates that there are roughly 18 million new cancer cases every year, predicted to increase to 21.5 million new cases by 2025. This leads to a requirement of analyzing roughly 200 million slides per year, assuming a conservative estimate of 10 whole slide images analyzed per patient over the course of their disease.  The results of these analyses will be employed both in research and in clinical care. The FDA has an active working group focusing on AI algorithms in pathology; this group is targeting TILs as a top priority clinical need.

Cancer registries and research organizations world-wide will make use of tumor immune characterizations. 
The National Cancer Institute Surveillance, Epidemiology, and End Results program (SEER) (\url{https://seer.cancer.gov/}) generates crucial epidemiological information for roughly 1/3 of all U.S. cancer patients. 
SEER currently supports Stony Brook University to develop algorithms and processes to support incorporation of TIL data into SEER cancer registries and reports (NCI CA225021). It is anticipated that TIL maps and specimen and patient level summaries generated from these maps will become crucial components of cancer epidemiological studies performed by NCI SEER and by the research community.  
At scale, the US SEER effort will involve analysis of TIL maps for roughly 10 million whole slide images on an annual basis. SEER data is generated locally at over 1,000 separate health care institutions.  
For logistical, legal, and contractual reasons, TIL data will also need to be generated locally.  
Local health care sites are frequently poorly resourced; efficient computation of TIL data is essential for adoption. Cancer working groups focused on TILs are notably international \cite{dieci2018update},\cite{morfouace2018transatlantic}; the need for inexpensive locally produced TIL predictions and resource constraints are similarly crucial in international settings. 

\subsubsection{Data Description}
\label{sec:performed_measured:data_description}

\begin{figure*}[ht!]
    \centering
    \includegraphics[width=1.0\textwidth]{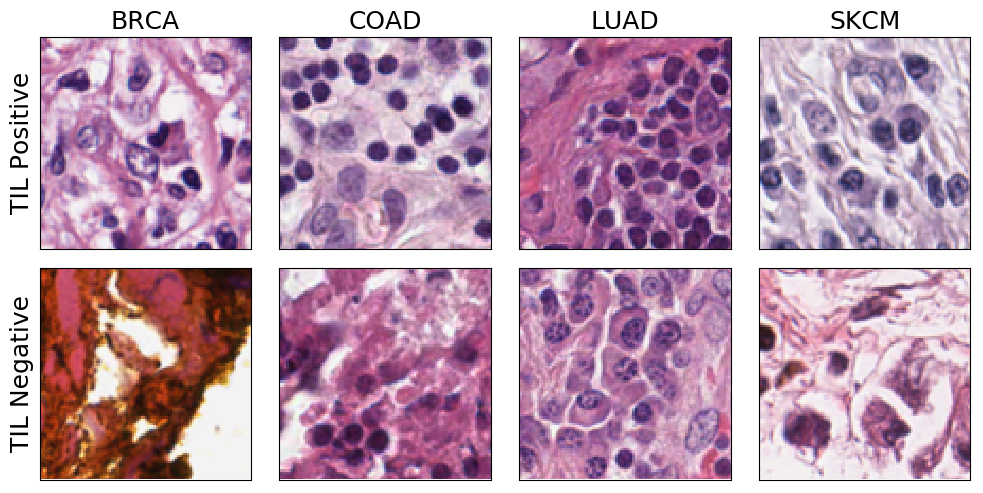}
    \caption{Examples from the dataset we use in this work.  We include positive and negative patches from four different cancer types: BRCA, COAD, LUAD, and SKCM.}
    \label{fig:tile_examples}
\end{figure*}

As the training set, we used 86,000 patches that were manually annotated with TIL classification \cite{saltz2018spatial}.  In the full set, there are 64,381 TIL negative patches and 21,773 TIL positive patches. All training patches used are in 100x100 pixel resolution, 20 times magnification, and are annotated as TIL positive or TIL negative. Examples of the training images and their labels are given in Figure \ref{fig:tile_examples}.  These training images represent seven different cancer types: invasive carcinoma of the breast (BRCA), colon adenocarcinoma (COAD), lung adenocarcinoma (LUAD), pancreatic adenocarcinoma (PAAD), prostate adenocarcinoma (PRAD), skin cutaneous melanoma (SKCM), and endometrial carcinoma of the uterine corpua (UCEC).  We use another 652 patches as our validation set (all from the LUAD cancer type), and 900 manually annotated patches from twelve  cancer types in total as the testing set.  The twelve cancer types are the seven listed above, as well as urothelial carcinoma of the bladder (BLCA), cervical squamous cell carcinoma and endocervical adenocarcinoma (CESC), lung squamous cell carcinoma (LUSC), rectal adenocarcinoma (READ), and stomach adenocarcinoma (STAD).

\begin{figure*}[ht]
    \centering
    \includegraphics[width=1.0\textwidth]{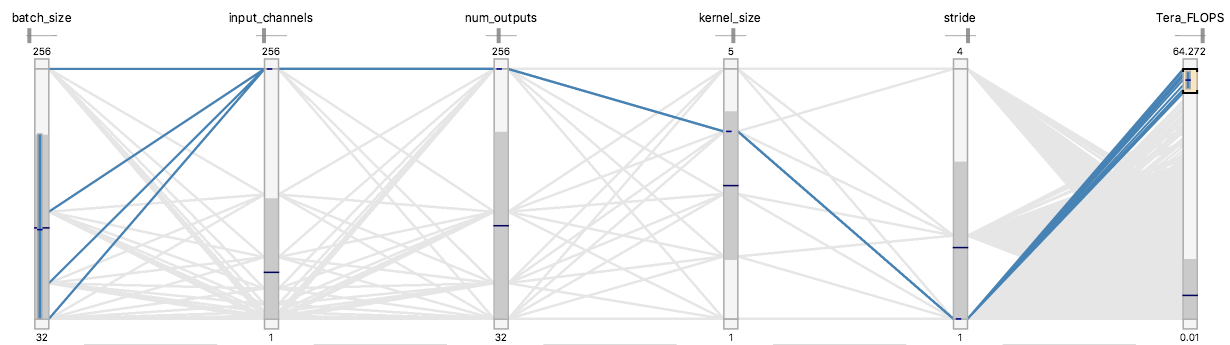}
    \caption{Parallel coordinates plot showing the results of a hyperparameter sweep for 10,000 convolution layers using tensor cores.  The top ten layer configurations are highlighted in blue.  All other parameter combination lines are shown in gray. Figure created using \cite{steed2013big}.}
    \label{fig:hyperparameter_sweep}
\end{figure*}

\subsection{System}

We performed our experiments using Oak Ridge National Laboratory's Summit supercomputer. Summit has 4,608 nodes, where each node contains two IBM POWER9 CPUs and six NVIDIA Volta GPUs, all of which are connected with NVIDIA’s high-speed NVLink. Each node has over half a terabyte of coherent memory (high bandwidth memory + DDR4) addressable by all CPUs and GPUs plus 800GB of non-volatile RAM that can be used as a burst buffer or as extended memory. The nodes are connected in a non-blocking fat-tree using a dual-rail Mellanox EDR InfiniBand interconnect\footnote{Summit description available here: \url{https://www.olcf.ornl.gov/olcf-resources/compute-systems/summit/}}. Each Volta GPU has tensor cores available that perform mixed-precision matrix multiply and accumulate calculations and account for a majority of Summit's compute capability.

\subsubsection{Measuring System Performance}

\begin{figure}
    \centering
    \includegraphics[width=\columnwidth]{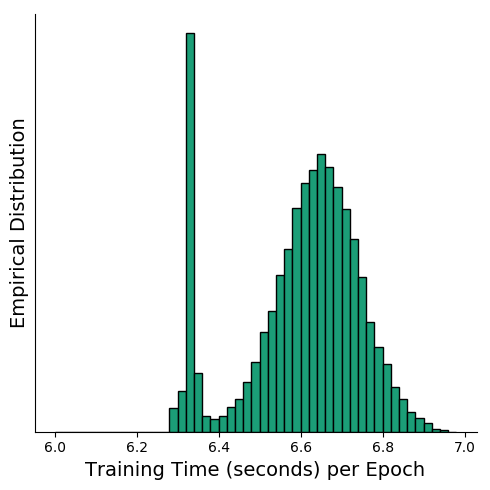}
    \caption{Distribution of average training time per epoch for a fixed network topology but varying learning algorithm parameters, measured on 4,585 distinct nodes on Summit (27,510 GPUs, one network per GPU).}
    \label{fig:timing_distribution}
\end{figure}

We measure a variety of performance characteristics of the deep learning networks generated by MENNDL, including power, energy, FLOP/s, and time for both training and prediction for each network evaluation.  We capture these values for a variety of reasons, particularly for use in our multi-objective optimization.  We instrument the training and prediction sections of the MENNDL code to capture these values.  We measure time using Python's time library, and we capture energy and power based on measurements from the NVIDIA System Management Interface (\texttt{nvidia-smi}) command line utility.

It is worth noting that measuring overall system performance for MENNDL has the additional difficulty that it is evaluating different network topologies and hyperparameter sets that have radically different performance characteristics, as shown in Figure \ref{fig:hyperparameter_sweep}. Thus, metrics such as FLOP counts and power usage must be measured per network.  However, even when the network topology and network hyperparameters are fixed, there are still differences in performance. When using a complex software stack such as MENNDL that includes third-party software (both closed and open source) and running it on a large-scale system such as Summit, it can also be difficult to determine why differences in performance can occur. Figure \ref{fig:timing_distribution} shows a distribution of measured training time per epoch when evaluating 27,510 networks;  the training operations per epoch for these networks should be very similar, if not identical, because they all have the same network topology and network hyperparameters.  As can be seen in the figure, there appear to be two separate distributions of training times, one centered around 6.35 seconds per epoch and the other around 6.7 seconds per epoch.  We have not yet been able to explain why there are two separate distributions, though we speculate that it could be attributed to a variety of software and/or hardware factors. However, these results indicate that, much as accuracy is often treated as a random variable when performing hyperparameter optimization, factors such as inference and training time should also be treated as random variables.

\section{Results}

\subsection{Application Results and Impact}

In this work, our primary focus was to use MENNDL to produce a network that can achieve comparable performance on the TIL identification task and achieve significantly faster prediction times than the previous state-of-the-art. In particular, the goal is to rapidly analyze existing cancer pathology data, such as that in the SEER program dataset (see Section \ref{sec:innovations_cancer}), and new data that is being added daily to the collection. 
To produce this network we used the new multi-objective optimization feature of MENNDL to maximize the F1-score and minimize prediction time.

Table \ref{tab:app_performance} shows how the approach in this work (MENNDL with multi-objective optimization to minimize prediction time, labeled as MENNDL in the table) compares with the baseline approach \cite{saltz2018spatial,hou2019sparse}, as well as an approach using an off-the-shelf network structure (Inception) that has been hand-tuned for this dataset.  In particular, Table \ref{tab:app_performance} reports the F1-score and the area under the curve (AUC) of the receiver operating characteristic (ROC) curve for each approach.  We use these measurements rather than accuracy on the dataset because the dataset is imbalanced.

\begin{table}[t]
\caption{Application Performance Comparison}
\label{tab:app_performance}
\centering
\begin{tabular}{|l|c|c|c|}
\hline
\textbf{Model Name} & \textbf{F1-Score} & \textbf{AUC} & \textbf{Prediction Rate} \\ \hline
Baseline \cite{saltz2018spatial,hou2019sparse} & 0.85 & 0.798 & Not reported \\ \hline
Inception & 0.87 & 0.899 & 433 patches/seconds \\ \hline
MENNDL (this work) & 0.83 & 0.839 & 7033 patches/second\\ \hline
\end{tabular}
\end{table}

\begin{figure}
    \centering
    \begin{tikzpicture}
    \node at (0, 4.5) {\includegraphics[width=.95\columnwidth, keepaspectratio]{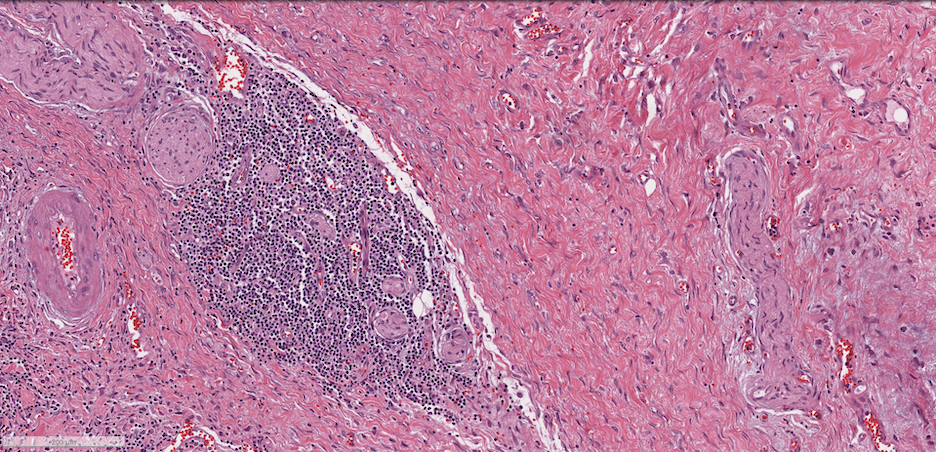}};
    
    \node at (0, 0.0)  {\includegraphics[width=.95\columnwidth, keepaspectratio]{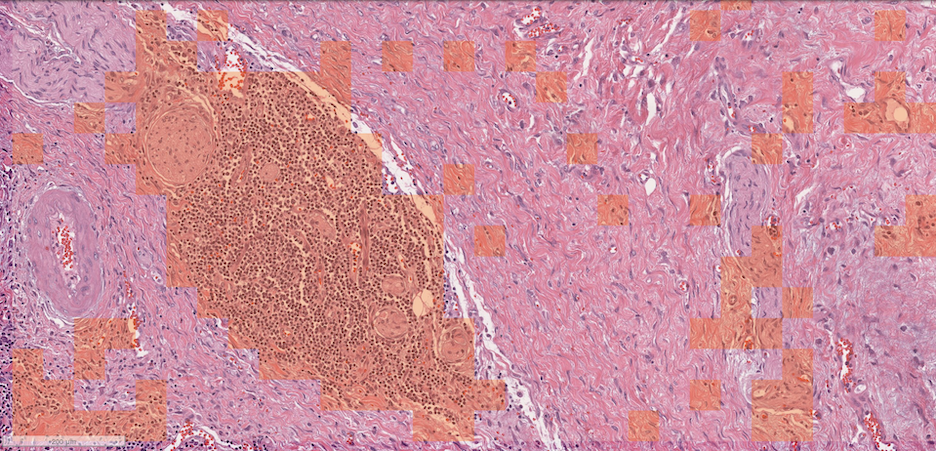}};
    \end{tikzpicture}
    \caption{A portion of a whole slide image from a PAAD example.  The top image is the original slide, and the bottom image highlights where TILs appear in the orange overlay, labeled using the MENNDL network.}
    \label{fig:whole_slide_image}
\end{figure}

As is noted in the table, the MENNDL network produces comparable F1-scores and AUC as the baseline and Inception results, but the \textbf{prediction times per patch are over $\mathbf{16\times}$ faster than the Inception results.}  Whole-slide images are on the order of 200,000 patches; in production use one would expect to run an ensemble of predictions using models produced by three independently generated training sets.  With the Inception network, using a single GPU (as a laboratory, pathologist, or researcher might have in their desktop computer) would take approximately 8 minutes to classify the TILs within an entire slide. The MENNDL-generated network achieves comparable accuracy, but can analyze a single whole slide image on a single GPU in roughly 30 seconds.  Assuming that three predictions are carried out, the MENNDL network would require 1.5 minutes and the Inception network 24 minutes.  Figure \ref{fig:whole_slide_image} shows a portion of a whole slide image with and without the TIL annotations that the MENNDL network produces. 

Slide scanners require 30 seconds to a few minutes to generate a whole slide image; the MENNDL network will have performance roughly matched to typical scanner data acquisition rates. Without the MENNDL network described here, the ability to collect data far outpaces the ability to analyze it effectively.  In terms of the SEER program dataset, as noted in Section \ref{sec:innovations_cancer}, at scale, we are interested in producing TIL maps for roughly 10 million whole slide images on an annual basis. As described in Section \ref{sec:innovations_cancer}, privacy and contractual issues inhibit aggregation of whole slide image data. Therefore, TIL maps will for the most part be generated locally by the thousands of institutions responsible for the generation of the SEER whole slide images. On a worldwide basis, national cancer research and surveillance organizations will have the same requirements for TIL maps and data products as the SEER program. Thus, one can expect that ultimately 100 million whole slide images will be analyzed per year world-wide to support these cancer epidemiology and surveillance programs, with a larger number generated for immediate clinical use.  For both the NCI SEER program and its international counterparts, there will be tremendous need for rapid, inexpensive local computation of TIL maps and TIL map data products. \textit{The MENNDL generated network described here will enable this need to be met by bringing the rate of image analysis up to speed with the rate of image collection.}   



\subsection{Scalability}

A secondary goal of the MENNDL tool is to be able to scale to use available HPC resources to quickly produce deep learning networks that can be practically deployed to non-HPC resources.  To demonstrate that MENNDL can utilize HPC resources effectively, we analyze MENNDL's performance on Summit, the world's fastest supercomputer. 

\begin{figure}[t]
    \centering
    \includegraphics[width=\columnwidth]{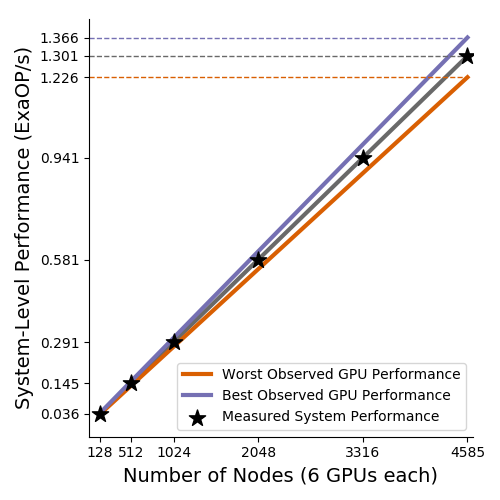}
    \caption{Weak scaling results.  Measured results are plotted in black/gray.  We also show the upper (purple) and lower (orange) bound on performance by showing the scaling performance assuming the best and worst observed network, respectively, was duplicated across all available GPUs and assuming that all GPUs perform identically.}
    \label{fig:weak}
\end{figure}

To determine our peak sustained performance, we measured on 4,585 nodes of Summit (the number of nodes available at the time of our test run), for a total of 27,510 GPUs. The measured value for our peak sustained performance is 1.301 exaflops, with an average of 47.286 teraflops per GPU,
which is approximately 38 percent of the theoretical max for Summit when using half precision arithmetic.

Figure \ref{fig:weak} shows our weak scaling results, in which we increase the number of networks to evaluate as we increase the number of nodes. 
We display the measured performance in Figure \ref{fig:weak}, but we also project the performance to give upper and lower bounds, assuming the benchmark network structure and hyperparamters given in Figure \ref{fig:wtn}, but varying the learning hyperparameters.  In particular, the upper bound is assuming that the learning hyperparameters that produced the most FLOP/s for that network are duplicated across all GPUs (and that all GPUs perform identically, which, as noted in Figure \ref{fig:timing_distribution}, may not be the case). Similarly, the lower bound is assuming that the learning hyperparameters that generated the fewest FLOP/s are duplicated.  In general, however, the variance across measured values is small, so MENNDL exhibits linear weak scaling in this case.  It is worth noting that if you vary the network architecture, then the upper and lower bounds can vary dramatically.

\section{Discussion and Conclusion}

There are important implications to cancer research coming from this work. Specifically, cancer researchers continue to study and develop new indicators to increase predictive measures of patient outcomes. For example, research shows that high densities of TILs correlate with favorable clinical outcomes \cite{Mlecnik2011}, such as overall survival and longer cancer-free periods for multiple cancer types \cite{ANGELL2013261}.   

The use of MENNDL and HPC provides a significantly faster way of quantitatively estimating the TIL density as well as characterization of TIL distribution patterns. This quantitative TIL characterization goes far beyond what a pathologist can estimate by inspecting tissue. The work presented here enables a high accuracy, high speed approach toward automating these TIL characterizations for millions of images. Thus, these images and their annotations can then be used to accelerate cancer research.  

The kind of network design optimization performed by MENNDL is enabled in a short amount of time only through the utilization of a high-performance computer such as Summit because of the size of the potential solution space size. Summit's GPU-centric heterogeneous design is ideal for quickly evaluating different configurations of deep learning networks. This resource provides ample compute capacity to create a population that can sufficiently explore the hyperparameter optimization space. Consequently, MENNDL running on HPC platforms creates a broader application impact than previous HPC-based deep learning approaches \cite{VanEssen:2015:LLB:2834892.2834897}\cite{Kurth:2018:EDL:3291656.3291724}.  Though we utilize HPC in this work to reach a good solution quickly, MENNDL can also be deployed on smaller systems, though the time-to-solution will be significantly increased.

In this work, MENNDL was significantly enhanced with the ability to optimize a deep learning network for multiple design objectives. Thus, not only can a network be optimized for accuracy on a given dataset, but also for other objectives such as prediction time or energy usage.  Training a deep learning network that has high accuracy on a given scientific dataset is just one step in the life cycle of a deep learning network.  In practice, the resulting network will then be used to analyze current data and future data as it is produced. Therefore, improvements in training time become much less significant than reductions in prediction times. Further, reductions in prediction time and improvements in computational performance provide the ability to analyze datasets that are currently beyond our capability to analyze. The ability to use HPC to produce networks that are capable of fast and accurate predictions makes HPC a significant enabling technology in using deep learning for scientific analysis. 

Although we have restricted our attention to the particular application of TIL identification in digital pathology images in this work, the MENNDL approach has much broader applicability.  MENNDL tailors the network design and hyperparameters of the network for a given dataset, automating and accelerating a process that is typically done manually by scientists for each individual dataset.  Pathology deep learning applications have been rapidly emerging; examples include pathology classification, tumor segmentation, and semantic segmentation of cell nuclei \cite{madabhushi2016image}, \cite{hou2016patch}, \cite{pantanowitz2018twenty}. We anticipate that MENNDL-generated networks are likely to be of great value for these applications. MENNDL has also been demonstrated successfully on a variety of applications, including datasets from high energy physics, small angle neutron scattering, medical imaging \cite{young2017evolving}, remote sensing \cite{Johnston:2017:OCN:3146347.3146352}, and scanning transmission electron microscopy data \cite{patton2018167}. Moving forward, MENNDL will ease the application of deep learning to a variety of other scientific datasets since it eliminates the burden from the scientist to hand-tune deep learning networks to operate on their data. 

\section*{Acknowledgements}
This material is based upon work supported by the U.S. Department of Energy, Office of Science, Office of Advanced Scientific Computing Research, Robinson Pino, program manager, under contract number DE-AC05-00OR22725 and by work supported by National Cancer Institute grants CA225021, CA180924 and CA215109.	 This research used resources of the Oak Ridge Leadership Computing Facility, which is a DOE Office of Science User Facility supported under Contract DE-AC05-00OR22725.

\bibliographystyle{IEEEtran}

\bibliography{references}

\end{document}